# DKP-AOM: results for OAEI 2015

Muhammad Fahad

*DISP Lab (http://www.disp-lab.fr), Université Lumière Lyon2*
*160 Boulevard de l'Université, Bron, FRANCE*
*firstname.lastname@univ-lyon2.fr*

**Abstract**

In this paper, we present the results obtained by our DKP-AOM system within the OAEI 2015 campaign. DKP-AOM is an ontology merging tool designed to merge heterogeneous ontologies. In OAEI, we have participated with its ontology mapping component which serves as a basic module capable of matching large scale ontologies before their merging. This is our first successful participation in the Conference, OA4QA and Anatomy track of OAEI. DKP-AOM is participating with two versions (DKP-AOM and DKP-AOM_lite), DKP-AOM performs coherence analysis. In OA4QA track, DKPAOM out-performed in the evaluation and generated accurate alignments allowed to answer all the queries of the evaluation. We can also see its competitive results for the conference track in the evaluation initiative among other reputed systems. In the anatomy track, it has produced alignments within an allocated time and appeared in the list of systems which produce coherent results. Finally, we discuss some future work towards the development of DKP-AOM.

*Keywords: Ontology matching, Ontology merging, disjoint knowledge, inconsistency, incompleteness, inconciseness, validation of mappings, verification of merged ontology*

## 1 Presentation of the System

Ontology merging is a process of building a new ontology from two or more existing ontologies with overlapping parts. The merged ontology can be either virtual or physical, but must be consistent, coherent and include all the information from the source ontologies [1]. Ontology merging is based on two primary steps. Firstly, the source ontologies are looked-up for correspondences between them. Secondly, duplicate-free and conflict-free union of source ontologies is achieved based on the established correspondences [2]. The first part mainly comes under the ontology matching, whereas the second part targets to achieve the merged ontology based on the results of the first part, i.e., mappings between source ontologies. To produce accurate merged ontology, there should be some mechanism to avoid erroneous intermediate mappings and also to merge them in such a way that produces consistent, complete and coherent merged ontology. There are many hurdles that come across in the generation of desired merged output. Firstly, ontological errors and design anomalies that can occur in the source ontologies detract from reasoning and inference mechanisms, and create bottleneck in their integration tasks [3]. In addition, conceptualization of domain, explication and modeling of knowledge over ontologies and semantic heterogeneities make their integration more difficult [4]. Secondly, even if the individual ontologies are free from errors, some of the identified mappings lead towards the erroneous situations producing several types of errors in the merged ontology [5]. For building an effective ontology merging algorithm, it is essential to incorporate ontological error checking during the validation of ontology mapping process and the verification of merged ontology to attain the accuracy of the resultant output.

In order to meet the above mentioned challenges for the ontology merging research, we proposed semi-automatic DKP-OM system implemented in Jena framework for the merging of heterogeneous ontologies with the intervention of a human user expert [6]. Later, we released a fully Automatic Ontology Merging (AOM) system named DKP-AOM implemented in OWLAPI 3.3 [7]. The name DKP comes from the concept of performing *Disjoint Knowledge Analysis (DKA)* and *Disjoint Knowledge Preservation (DKP)* during the merging process. *Disjoint Knowledge Analysis* plays a vital role in controlling the search space for finding similarities between source ontologies. Look-up within disjoint partitions of source ontologies significantly reduces the time complexity of the mapping phase. *Disjoint Knowledge Preservation* in the merged ontology helps to preserve disjoint axioms in the sub-hierarchies of merged ontology to avoid incompleteness in the resultant merged ontology. In this way, it also pin-points different conflicts between source ontologies based on disjoint axioms in the source ontologies and detects inconsistent mappings. Computed mappings that lead in many cases to a large number of unsatisfiable classes are eliminated so that the resultant merged ontology should not suffer from inconsistencies. The next sub-sections provide more details about DKP-AOM and then discuss our results of OAEI participation.

## 1.1 State, purpose, general statement

Our system DKP-AOM follows a five step methodology as illustrated in the Figure 1. First, it generates the intermediate models (OWL-DL Graphs) of source ontologies and does preprocessing on the concept URIs and labels. Second using these graphs, *MatchManager* component performs the first level task of finding the initial lexical, synonym and axiomatic based mappings between concepts. For this, it first builds the search space based on disjoint axioms present inside the source ontologies for finding the correspondences between the ontologies.

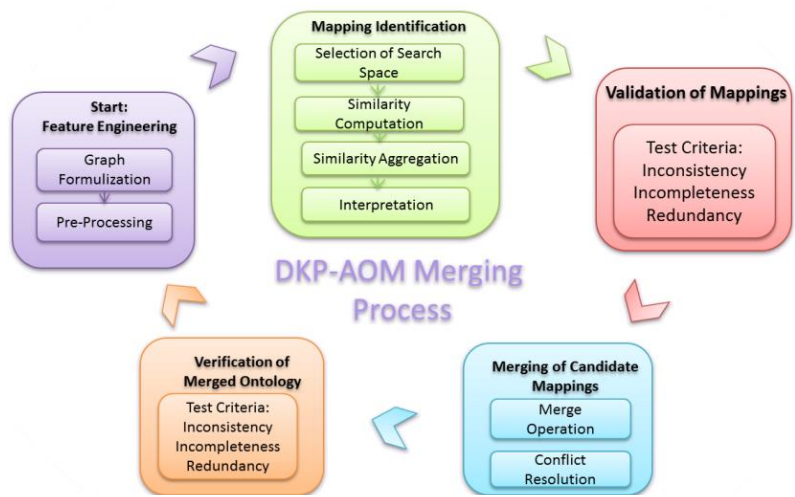

Figure 1. DKP-AOM merging process

*MatchManager* performs the following similarity measures to detect mappings:

**Concept Similarity.** It employs several basic matchers to find mappings between concepts. Concept URI similarity ($Sim_{uri}$) and Label similarity ($Sim_{lab}$) computes lexical and synonym based correspondences at the element level between source ontologies. Lexical similarity finds the string-based correspondences based on *SimMetric*[*]. Synonym similarity is computed based on the lexical database *Wordnet* that helps to detect the concepts which have the same meanings but are lexically different.

**Datatype and Object property Similarity.** Correspondences between Datatype and Object properties are identified on the basis of their URI and Label similarities. It also considers domain and range assosicated with them to detect a perfect match.

**Inheritance based Similarity.** It also considers that an inheritance is a vital factor to detect the mapping candidates between source ontologies. It increases a level of confidence that the detected mappings have not just a lexical similarity, but a real mapping having parent-child relation as well. *Inheritance* matching is done after the *Concept label* similarity and *synonym* similarity. Consider a scenario, where an ontology O1 contains Person and PhD concepts. Person is defined as: *{Person subclassof hasName some String},* and concept PhD is defined as a subconcept of Person concept as: *{PhD subClass Person}.* In ontology O2, there is a PhD candidate that is defined as: *{PhD subclass of hasName some String}.* In such a case we get the basic mappings between O1:PhD, and O2:PhD. Then the inheritance matcher plays an important role by matching the inheritance of the restriction from Person to PhD from O1 with the restriction in O2 and adds the confidence level of their similarity. In this way, an inheritance matching has a potential impact in the proposed solution. This similarity will help for the detection of axiomatic similarities between concepts.

**Concept DL Axiomatic Similarity ($Sim_{axm}$).** OWL classes are described through the class descriptions/expressions that enrich the background information of the concepts and represent the constraints of real world situations. For finding the accurate semantic similarity between the concepts of ontologies, DL axioms can help significantly as they define the context of the concepts. They link the concept by different means that depict the concept's real semantics. Therefore, it gets axiomatic definitions, which can be formed from the union, complement, intersection and restriction operators applied on the primitive concept or the anonymous concept and/or by their boolean combinations, and performs matching to detect such DL axiom similarities. This is the most difficult part of matching, although most significant as well. Most ontologies in *Conference track* in OAEI were equipped with OWL DL axioms of various kinds, therefore it opens a way to use semantic matching.

*MatchManager* aggregates the individual similarities between ontologies and propagates the results to *ConsistencyChecker* for the validation of mappings. Third, *ConsistencyChecker* has many detectors that make the validation of each mapping found in the initial stage so that the merged ontology stays consistent with reference to the source ontologies. When the initial mappings pass the consistency test, *ConsistencyChecker* passes the mappings to the *Reasoner*.

Fourth, *Reasoner* aggregates the output of different similarity measures, resolves conflicts and merges mappings to generate a global merged ontology. This *Reasoner* is a component of DKP-AOM system and not an open source DL Reasoner engine. It implements various patterns (see details in [7]) for the automatic merging of source ontologies in case of different types of conflicts and structural differences in source ontologies.

Finally, it compiles the output as a merged global ontology automatically or a final list of consistent mappings as required by the end user. Our merging algorithm imports the first ontology as the merged ontology and then performs several operations to build the combined definitions for each of the concepts from the source ontologies. Each of the axiomatic definitions from the source ontologies are matched together, merging is performed on them, and the *combined rich axioms* are added in the merged ontology. Our merging algorithm performs deletion of axioms or the rewriting of some of them in order to preserve desired consequences while removing the undesired ones. Merging of axiomatic definitions really achieves a richer merged ontology which captures sufficient definitions from the source ontologies. Finally, it applies the quality criteria and ensures the ultimate goal of achieving the satisfiability of merged ontology by checking the correctness and consistency of concepts, properties, and axioms of the generated ontology.

---

[*] http://sourceforge.net/projects/simmetrics

## 1.2 Specific Techniques Used

**Data Preprocessing:** Linguistic analysis of concept labels and properties is done with the help of MorphAdorner† (version 1.0). MorphAdorner is helpful in various cases especially the lemmatization process is worth useful for detecting the base words of terms and irregular verbs used in source ontologies. For example, concept "students" to lemma "student" and properties ("Accepted", "Accepting", "Accept", and "Accepts") to their base "accept". MorphAdorner is really helpful for the detection of similarities between properties which are usually not in the base form.

**Search Space Analysis based on Disjoint Partitions:** It is very important to build the search space for the lookup of mappings between ontologies. In general, it requires exhaustive analysis (or complete comparison) for the similarity computation between the concepts of ontologies, where each concept *c* of the ontology $O_a$ is matched with each concept *c'* of the ontology $O_b$. The restriction of look-up within disjoint partitions minimizes the search space for the mapping computation [8]. For an example, consider conference ontologies illustrated in Figure 2. There are 14 concepts in *O1:CRS_DR* ontology and 36 concepts in *O2:CMT ontology*. In *CRS_DR* ontology, three disjoint axioms between level-1 concepts, partition the domain concepts in four non-overlapping domains, i.e., *Program, Person, Document and Event*. *CMT* ontology partitions the concept into six disjoint categories, i.e., *Person, Decision, Document, Conference, Preference*, etc. These ontologies allow the concept mapping search space look-up within disjoint partitions. For example, search spaces look-up within (*O1:Person, O2:Person*), (*O1:Document, O2:Document*), (*O1:Event, O2:Conference*). For concept matching, it needs 14*36 = 504 comparisons. But, lookup in disjoint partitions makes the search space much smaller, and requires maximum 155 comparisons (by manual calculation) for mapping the concepts of these ontologies.

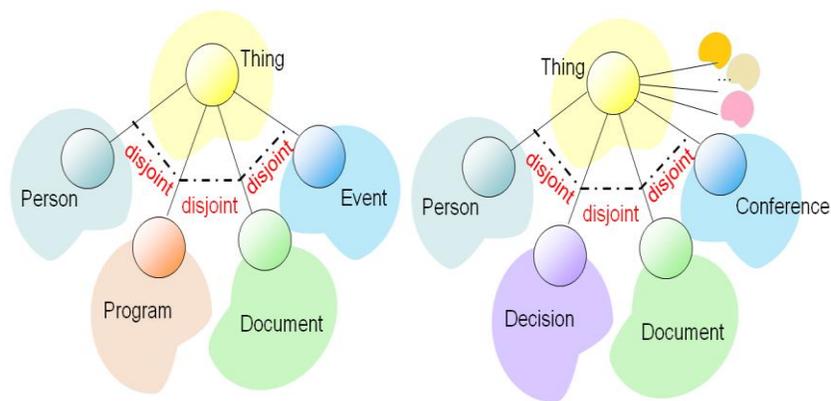

Figure 2. Top level Disjoint Partitions in CRS_DR and CMT conference ontologies

† http://morphadorner.northwestern.edu

We believe this idea of *Divide and Conquer* is helpful for the large ontologies. The more disjoint axioms are modeled in the input ontologies, the more search space is restricted, which impacts in lower number of comparisons for the identification of mappings. We call this disjoint based partitioning strategy as *Divide and Conquer* approach. In fact, it resembles, but not achieves a full or successful conquer in all scenarios of partitioning, as divide and conquer strategies found in an algorithm domain where the divided partition is at last conquered with success. For example, in case of failure, when a concept does not found its mapping concept in the divided partition, it is matched with the other level of concepts to find its mapping (just like exhaustive search in the entire space but step-by-step looking in sibling hierarchy and then search upward).

**Validation of Mappings:** For the evaluation of ontologies, Gomez-Perez constructed an error taxonomy as a guideline for ontology engineers to help building well-formed and well-structured classification of concepts in the ontologies. She defined three classes of ontological errors that might occur when modeling the conceptualization into taxonomies, i.e., *Inconsistency, Incompleteness and Redundancy* [9] as illustrated in Figure 3. Inconsistency in ontology means that there is some *sort of contradictory knowledge* inferred from the concepts, definitions and instances within the ontology. It creates ambiguity, contradictions in interpretations and compromises precision of results. Incompleteness occurs when ontologists model the domain knowledge in the form of concepts, properties and definitions, but *overlooked some of the important information* about the domain. The incompleteness of domain knowledge lacks reasoning and prevents inference mechanisms. The other important task is to make ontologies concise *without repeating and replicating same information* so that they store only necessary and sufficient knowledge about the concepts, axioms and properties. Redundancy errors not only compromise conciseness and usability, but also create problems for the maintenance and manageability of ontologies. We used this framework as a test criteria for the validation of mapping and verification of merged ontology [10]. Computed initial mappings that lead to unsatisfiable classes in merged ontology are eliminated so that the resultant merged ontology should not suffer from inconsistencies. This test criteria serves best for the detection of inconsistent mappings and also for ensuring the satisfiability of a merged ontology. This is one of the reasons that our system has produced alignments which allowed to answer **all the queries** of the OA4QA track evaluation.

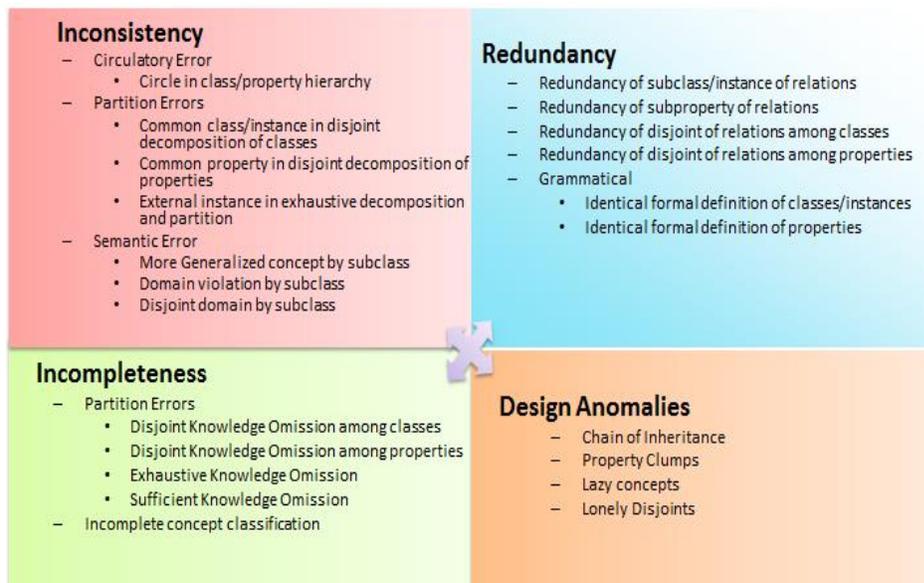

Figure 3. Test criteria for the validation of mappings and verification of merged ontologies

## 1.3 Adaptations made for the evaluation

As you read above, DKP is an automatically merging system. Therefore it was developed based on user GUIs such as source ontology trees for display, visual alignments between ontologies, merged ontology tree, etc. The original version of DKP has changed and these visual components are removed so that it can participate under the seals platform. However, still it needs proper clean-up to improve its runtime for the future OAEI participations.

## 1.4 Link to the system

Various versions of DKP can be found at my personal site: http://sites.google.com/site/mhdfahad under *plugins* tab. The mapping system is separated from the merging system, and can be downloaded according to needs. For the merging of ontologies, use the same command of seals platform with –o following three paths, two for source ontologies and one for the output merged ontology. As a result of this command, a list of ontology mappings and a resultant merged ontology are produced.

# 2 Results

In order to show the efficiency and effectiveness of our system, this year we participated in Conference, OA4QA and Anatomy tracks. The results are very encouraging provided by the OAEI 2015 campaign as our system is acceptable and comparable with other participants, and are discussed in the following subsections.

## 2.1 Conference

The goal of conference track is to find alignments among 16 ontologies relatively smaller in size (between 14 and 140 entities) but rich in semantic heterogeneities about the conference organization domain. As a result, Alignments are evaluated automatically against reference alignments. Therefore, it is very interesting to measure the Precision, Recall and F-measure of our system and also does a comparison between existing systems to see their performance on real world datasets. Table 1 presents the results obtained by running DKP-AOM on the Conference track of OAEI campaign 2015. Our system DKP-AOM has produced very competitive results among top ranked systems. Our precision measure is significantly high, recall is good giving comparable F-measure value to depict a real effort towards detecting heterogeneities for the goal of ontology matching.

| Matcher | Runtime | Precision | F-Measure | Recall |
|---|---|---|---|---|
| DKP-AOM | 9913 | 0.844 | 0.626 | 0.498 |

Table 1. DKP-AOM results on conference track ontologies

## 2.2 Ontology Alignment for Query Answering (OA4QA)

In this track, our system has out-performed and generated excellent results (see Table 2). Precision and recall has calculated with respect to the ability of the generated alignments to answer a set of queries in an ontology-based data access scenario where several ontologies exist. AML, DKP-AOM, LogMap, LogMapC and XMap were the only matchers whose alignments allowed to answer **all the queries** of the evaluation. The best global results have been achieved for violations queries, that has been correctly covered by AML, DKP-AOM, LogMap, COMMAND, LogMapC and XMAP, in decreasing order of f-measure w.r.t. RA1. Notably, our DKP-AOM achieved an impressive *f-measure* of 0.999 w.r.t. RAR1, showing an effective handling of logical violations.

| Successful queries | Precision (RA1) | Recall (RA1) | F-Measure (RA1) | Precision (RAR1) | Recall (RAR1) | F-Measure (RAR1) |
|---|---|---|---|---|---|---|
| **Global Evaluation Results** | | | | | | |
| 18/18 | 0.667 | 0.618 | 0.635 | 0.666 | 0.639 | 0.648 |
| **Advanced Queries Results** | | | | | | |
| 5/5 | 0.200 | 0.100 | 0.133 | 0.200 | 0.100 | 0.133 |
| **Basic Queries Results** | | | | | | |
| 6/6 | 0.667 | 0.667 | 0.667 | 0.667 | 0.667 | 0.667 |
| **Violations Queries Results** | | | | | | |
| 7/7 | 1.000 | 0.947 | 0.967 | 0.999 | 1.000 | 0.999 |

Table 2. DKP-AOM results on Ontology Alignment for Query Answering (OA4QA) track

## 2.3 Anatomy

The anatomy real world case is about matching two very large biomedical ontologies, i.e., Adult Mouse Anatomy (2744 classes) and the NCI Thesaurus (3304 classes) describing the human anatomy. We participated with two versions DKP-AOM and DKP-AOM_lite, DKP-AOM performs coherence analysis. The evaluation was run on a server with 3.46 GHz (6 cores) and 8GB RAM, with allocated time less than an hour. This year 2015, there are 11 different systems (not counting different versions) which generated an alignment, out of them only four systems participated in the anatomy track for the first time. These are COMMAND, GMap, JarvisOM and DKP-AOM (with two version). Two of them COMMAND and GMap run out of memory and could not finish execution with the allocated amount of memory, therefore, their execution times are not fully comparable to the other systems. Our systems have produced results within an allocated time, illustrated in Table 3 with other systems.

Importantly, our DKP-AOM achieves coherency and became in the list of 7 systems which produced only coherent mappings. It has also generated only trivial correspondences.

F-Measure of DKP-AOM_lite (0.763) is very near to the baseline which is based on (normalized) string equivalence (StringEquiv, 0.766), with difference of only .003.

| Matcher | Runtime | Size | Precision | F-Measure | Recall | Recall+ | Coherent |
|---|---|---|---|---|---|---|---|
| COMMAND | 63127* | 150 | 0.293 | 0.053 | 0.029 | 0.042 | x |
| GMap | 2362** | 1344 | 0.916 | 0.861 | 0.812 | 0.534 | -- |
| JarvisOM | 217 | 458 | 0.365 | 0.169 | 0.11 | 0.01 | - |
| DKP-AOM | 370 | 201 | 0.995 | 0.233 | 0.132 | 0.0 | x |
| DKP-AOM_lite | 476 | 949 | 0.991 | 0.763 | 0.62 | 0.042 | - |
| StringEquiv | - | 946 | 0.997 | 0.766 | 0.622 | 0.000 | - |

Table 3. Results of first time participating systems on Anatomy track

## 3   Conclusion and Future Directions

The participation of DKP-AOM in OAEI 2015 is a success in the conference and OA4QA track. In OA4QA track, DKPAOM out-performed in the evaluation and generated accurate alignments which allowed to answer all the queries of the evaluation. In Conference track, it produced competitive results among top ranked systems. It also presented intermediate results in the Anatomy track and comes in the list of 7 matching system which produce coherent results. However, the whole framework of DKP-AOM is very huge and the participated version needs more effort of development to achieve more success in the upcoming OAEI. We plan to integrate synonym based mappings in the participated version. In addition, we plan to implement all the test criteria inside the DKP-AOM and present it as a complete system that achieves consistency, completeness and coherency.

Our technique of building search space is based on the disjoint partitions available in the source ontologies (that are very rarely present in the dataset ontologies). One of our future directions is to devise a disjoint learning algorithm to identify and make disjoint partitions automatically in the source ontologies, even if disjoint partitions were not present in the source ontologies before their merging.

## References

1. Bruijn, J.d., Ehrig, M., Feier, C., Martín-Recuerda, F., Scharffe, F., and Weiten., M., Ontology mediation, merging and aligning. In Semantic Web Technologies. Wiley 2006
2. Euzenat, J., and Shvaiko, P., Ontology Matching. Springer, 2007, ISBN 978-3-540-49611-3.
3. Fahad, M., Qadir, M.A., Noshairwan, M.W., Ontological Errors - Inconsistency, Incompleteness and Redundancy. In Proceedings of 10th Intl Conference on Enterprise Information Systems, pp. 253-285, 2008, Spain, Springer,
4. Klein, M., (2001): Combining and relating ontologies: an analysis of problems and solution. In Proc. of Workshop on Ontologies and Information Sharing (IJCAI), pp. 53-62. Seattle, USA (2001)
5. Fahad, M., and Qadir, M.A., A Framework for Ontology Evaluation, 16th ICCS Supplement Proceeding, vol. 354, 2008, France, pp.149-158.
6. Fahad, M., Qadir, M.A., Noshairwan, W., Iftakhir, N., DKP-OM: A Semantic based Ontology Merger, Proceedings of 3rd International Conference on Semantic Technologies (I-Semantics 07) Graz, Austria, 2007, Pages 313-322
7. Fahad, M., Moalla, N., Bouras, A., Detection and Resolution of Semantic Inconsistency and Redundancy in an Automatic Ontology Merging System, Journal of Intelligent Information System (JIIS), Vol. 39(2) pp. 535-557, 29/4/2012, DOI 10.1007/s10844-012-0202-y
8. Fahad, M., Moalla, N., Bouras, A., Qadir, M.A., Farukh, M., Disjoint Knowledge Analysis and Preservation in Ontology Merging Process, proceedings of 5th International Conference on Software Engineering Advances (ICSEA'10), IEEE CS, August 22-27, 2010 - Nice, France.
9. Gómez-Pérez, A., (2001): Evaluating ontologies: Cases of Study. IEEE Intelligent Systems and their Applications, vol. 16(3): 391–409, (2001)
10. Fahad, M., Moalla, N., Bouras, A., Towards ensuring Satisfiability of Merged Ontology, International conference on computational science, ICCS 2011, Procedia Computer Science 4 (2011), pp. 2216–222, 1-3 june, 2011